\pdfoutput=1

\documentclass[]{article}
\usepackage{graphicx}
\usepackage{algorithmicx}
\usepackage{algorithm}
\usepackage[noend]{algpseudocode}
\usepackage{amsmath}
\usepackage{amssymb}
\usepackage{color}
\usepackage{enumitem}
\usepackage[comma,authoryear]{natbib}
\usepackage{proceed2e}
\usepackage{subcaption}
\usepackage{url}
\usepackage[]{hyperref}

\title{Efficient Inference of\\ Gaussian Process Modulated Renewal Processes
 \\with Application to Medical Event Data}

\author{ {\bf Thomas A.~Lasko} \\
Computational Medicine Laboratory, Department of Biomedical Informatics \\
Vanderbilt University School of Medicine \\
Nashville, TN 37203 \\
}

\newcommand{\figref}[1]{(Figure \ref{fig:#1})}
\newcommand{\eqnref}[1]{\eqref{eq:#1}}
\newcommand{\secref}[1]{(Section \ref{sec:#1})}

\newcommand{\GP}{\ensuremath{\mathcal{GP}}}

\DeclareMathOperator{\Prob}{\mathbf P}
\DeclareMathOperator{\Exp}{Exp}
\DeclareMathOperator{\Unif}{Unif}

\begin{document}

\maketitle

\begin{abstract}
  The episodic, irregular and asynchronous nature of medical data render them difficult substrates
  for standard machine learning algorithms. We would like to abstract away this difficulty for the
  class of time-stamped categorical variables (or \emph{events}) by modeling them as a renewal
  process and inferring a probability density over continuous, longitudinal, nonparametric intensity
  functions modulating that process.  Several methods exist for inferring such a density over
  intensity functions, but either their constraints and assumptions prevent their use with our
  potentially bursty event streams, or their time complexity renders their use intractable on our
  long-duration observations of high-resolution events, or both. In this paper we present a new and
  efficient method for inferring a distribution over intensity functions that uses direct numeric
  integration and smooth interpolation over Gaussian processes. We demonstrate that our direct
  method is up to twice as accurate and two orders of magnitude more efficient than the best
  existing method (thinning). Importantly, the direct method can infer intensity functions over the
  full range of bursty to memoryless to regular events, which thinning and many other methods
  cannot. Finally, we apply the method to clinical event data and demonstrate the face-validity of
  the abstraction, which is now amenable to standard learning algorithms.

\end{abstract}

\section{Introduction}
One of the hurdles for identifying clinically meaningful patterns in medical data is the fact that
much of that data is sparsely, irregularly, and asynchronously observed, rendering it a poor
substrate for many pattern recognition algorithms.

A large class of this problematic data in medical records is time-stamped categorical data such as
billing codes. An ICD-9 billing code, for example, with value 714.0 (Rheumatoid Arthritis) gets
attached to a patient record every time the patient makes contact with the healthcare system for a
problem or activity related to her arthritis. This could be an outpatient doctor visit, a laboratory
test, a physical therapy visit, the discharge event of an inpatient stay, or any other billable
event. These events occur at times that are in general independent from events for other conditions.

We would like to learn things from the patterns of these clinical contact events both within and
between diseases, but their (often sparse and) irregular nature makes it difficult to apply standard
learning algorithms to them. To abstract away this problem, we consider the data as streams of
events, one stream per code or other categorical label. We model each stream as a modulated renewal
process and use the process's modulation function as the abstract representation of the continuous,
longitudinal intensity of the patient's contact with the healthcare system for a particular problem
at any point in time. The resulting inference problem is to estimate a probability density over the
renewal process parameters and intensity functions given the raw event data.

The practical utility of using a continuous function density to couple standard learning algorithms
to sparse and irregularly observed continuous variables has been previously demonstrated
\citep{Lasko2013}. Unfortunately, the method of inferring such densities for continuous variables is
not applicable to categorical variables. This paper presents a method that achieves the inference
for categorical variables.

Our method models the log intensity functions non-parametrically as Gaussian processes, and uses
Markov Chain Monte Carlo (MCMC) to infer a posterior distribution over intensity functions and model
parameters given the events \secref{mrp}.

There are several existing approaches to making this inference \secref{prior-work}, but all of the
approaches we found have either flexibility or scalability problems with our clinical data. For
example, clinical event streams can be bursty, and some existing methods are unable to adapt to or
adequately represent this.

In this paper we demonstrate using synthetic data that our approach has accuracy, efficiency, and
flexibility advantages over the best existing method \secref{experiment-synthetic}. We further
demonstrate these properties using synthetic data that mimics our clinical data, under conditions
that no existing method that we know of is able to satisfactorily operate
\secref{experiment-synthetic}. Finally, we use our method to infer continuous abstractions over
real clinical data \secref{experiment-clinical}.

\section{Modulated Renewal Process Event Model}
\label{sec:mrp}

A renewal process models random events by assuming that the interevent intervals are independent and
identically distributed (iid). A modulated renewal process model drops the iid assumption and adds a
longitudinal intensity function that modulates the event rate with respect to time.

We consider a set of event times $T = \{t_0, t_1, \dotsc, t_n\}$ to form an event stream that can be
modeled by a modulated renewal process. For this work we choose a modulated gamma
process \citep{Berman1981}, which models the times $T$ as
\begin{multline}
  \label{eq:modulated-gamma}
  \Prob(T ; a, \lambda(t)) = \\
  \frac{1}{\Gamma(a)^n}\prod_{i = 1}^n \lambda(t_i)(\Lambda(t_i) - \Lambda(t_{i-1}))^{a-1}e^{-\Lambda(t_n)},
\end{multline}
where $\Gamma(\cdot)$ is the gamma function, $a > 0$ is the shape parameter, $\lambda(t) > 0$ is the
modulating intensity function, and $\Lambda(t) = \int_0^t \lambda(u)\, du$.

Equation \eqnref{modulated-gamma} is a generalization of the homogeneous gamma process $\gamma(a,
b)$, which models the interevent intervals ${x_i = t_i - t_{i - 1}}, {i = 1 \dotsc n}$ as positive
iid random variables:
\begin{equation}
\label{eq:gamma}
\gamma(x|a, b) = \Prob(x;a, b) = \frac{1}{\Gamma(a)b^a}x^{a - 1}e^{-x/b},
\end{equation}
where $b$ takes the place of a now-constant $1/ \lambda(t)$, and can be thought of as the time scale
of event arrivals.

The intuition behind \eqnref{modulated-gamma} is that the function $\Lambda(t)$ warps the event
times $t_i$ into a new space where their interevent intervals become draws from the homogeneous
gamma process of \eqnref{gamma}. That is, the warped intervals $w_i = \Lambda(t_i) -
\Lambda(t_{i-1})$ are modeled by $w_i \sim \gamma(a, b)$.

For our purposes, a gamma process is better than the simpler and more common Poisson process because
a gamma process allows us to model the relationship between neighboring events, instead of assuming
them to be independent or memoryless. Specifically, parameterizing $a < 1$ models a bursty process,
$a > 1$ models a more regular or refractory, and $a = 1$ produces the memoryless Poisson
process. Clinical event streams can behave anywhere from highly bursty to highly regular.

We model the log intensity function ${\log \lambda(t) = f(t) \sim \GP(0, C)}$ as a draw from a Gaussian
process prior with zero mean and the squared exponential covariance function
\begin{equation}
  \label{eq:cov}
  C(t_i, t_j) = \sigma e^{-\bigl(\frac{t_i - t_j}{l}\bigr)^2},
\end{equation}
where $\sigma$ sets the magnitude scale and $l$ sets the time scale of the Gaussian process. We
choose the squared exponential because of its smoothness guarantees that are relied upon by our
inference algorithm, but other covariance functions could be used.

In our application the observation period generally starts at $t_{\min} < t_0$, and ends at
$t_{\max} > t_n$, and no events occur at these endpoints. Consequently, we must add terms to
\eqnref{modulated-gamma} to account for these partially observed intervals. For efficiency in
inference, we estimate the probabilities of these intervals by assuming that $w_0$ and $w_{n+1}$ are
drawn from a homogeneous $\gamma(1,1)$ process in the warped space. The probability of the leading
interval $w_0 = \Lambda(t_0) - \Lambda(t_{\min})$ is then approximated by $\Prob(w \ge w_0) =
\int_{w_0}^\infty e^{-w}\,dw = e^{-w_0}$, which is equivalent to $w_0 \sim \gamma(1,1)$. The trailing interval is treated similarly.

Our full generative model is as follows:
\begin{enumerate}[noitemsep]
\item \label{item:hyperpriors} $l \sim \Exp(\alpha)$ \\
  $\log \sigma \sim \Unif (\log\sigma_{\min},  \log \sigma_{\max})$ \\
  $\log a \sim  \Unif(\log a_{\min}, \log a_{\max})$\\
  $b = 1$
\item \label{item:GP}$f(t) \sim \GP(0, C)$ using \eqnref{cov}
\item \label{item:transform}$\lambda(t) = e^{f(t)}$
\item \label{item:integral} $\Lambda(t) = \int_0^t \lambda(u)\, du$
\item $w_0 \sim \gamma(1,1)$; $w_{i > 0} \sim \gamma(a, b)$
\item \label{item:inversion} $t_i = t_{\min} +\Lambda^{-1}(\sum_{j=0}^i w_j)$

\end{enumerate}

Step \ref{item:hyperpriors} places a prior on $l$ that prefers smaller values, and uninformative
priors on $a$ and $\sigma$. We set $b = 1$ to avoid an identifiability problem. (\citet{Rao2011}
set $b = 1/a$ to avoid this problem. While that setting has some desirable properties, we've found
that setting $b = 1$ avoids more degenerate solutions at inference time.)

\subsection{Inference} Given a set of times $T$, we use MCMC to simultaneously infer posterior
distributions over the intensity function $\lambda(T)$ and the parameters $a$, $\sigma$, and $l$
(Algorithm \ref{alg:inference}), where for simplicity we denote $\lambda(T) = \{\lambda(t):t \in
T\}$. On each round we first use slice sampling with surrogate data \citep[][code publicly
available]{Murray2010a} to compute new draws of $f(t)$, $\sigma$, and $\l$ using
\eqnref{modulated-gamma} as the likelihood function (with additional factors for the incomplete
intervals at the ends). We then sample the gamma shape parameter $a$ with Metropolis-Hastings moves
under the same likelihood function.

One challenge of this direct inference is that it requires integrating $\Lambda(t) = \int_0^t
\lambda(u)\, du$, which is difficult because $\lambda(t)$ does not have an explicit expression. Under
certain conditions, the integral of a Gaussian process has a closed form \citep{Rasmussen2003}, but
we know of no closed form for the integral of a log Gaussian process. Instead, we compute the
integral numerically, relying on the smoothness guarantees provided by the covariance function
\eqref{eq:cov} to provide high accuracy.

The efficiency bottleneck of the update is the $O(m^3)$ complexity of updating the Gaussian process
$f$ at $m$ locations, due to a matrix inversion. Naively, we would compute $f$ at all $n$ of the
observed $t_i$, with additional points as needed for accuracy of the integral. To improve
efficiency, we do not directly update $f$ at the $t_i$, but instead at $k$ uniformly spaced points
\mbox{$\hat{T} = \{\hat{t}_j = t_{\min} + jd$\}}, where \mbox{$d = \frac{t_{\max} -
    t_{\min}}{k-1}$}. We then interpolate the values $f(T)$ from the values of $f(\hat{T})$ as
needed. We set the number of points $k$ by the accuracy required for the integral. This is driven by
our estimate of the smallest likely Gaussian process time scale $l_{\min}$, at which truncate the
prior on $l$ to guarantee $d \ll \l_{\min} \le l$. The efficiency of the resulting update is $O(k^3)
+ O(n)$, with $k$ depending only on the ratio $l_{\min} / (t_{\max} - t_{\min})$.

It helps that the time constant driving $k$ is the scale of changes in the modulating intensity
function instead of the scale of interevent intervals, which is usually much smaller. In practice,
we've found $k = 200$ to work well for nearly all of our medical data examples, regardless of the
observation time span, resulting in an update that is linear in the number of observed points.

Additionally, the regular spacing in $\hat{T}$ means that its covariance matrix generated by
\eqnref{cov} is a symmetric positive definite Toeplitz matrix, which can be inverted or solved in a
compact representation as fast as $O(k \log^2 k)$ \citep{Martinsson2005}. We did not include this
extra efficiency in our implementation, however.

\begin{algorithm}[hbt]
  \caption{Intensity Function and Parameter Update}
  \label{alg:inference}
  \begin{algorithmic}[1]
    \Require Event times $T$, regular grid $\hat{T}$, current function $f$ and parameters $\sigma$, $l$, and $a$
    \Ensure Updated $f$, $\lambda$, $\Lambda$, $\sigma$, $l$, and $a$ with likelihood $p$
    \State Update $f(\hat{T}), \sigma, l$, using slice sampling
    \State  Compute $f(T)$ by smooth interpolation of $f(\hat{T})$
    \State $\lambda({T \cup \hat{T}}) \gets e^{f(T \cup \hat{T})}$
    \State Compute $\Lambda(T)$ from $\lambda({\hat{T}})$ numerically
    \State Compute $p$ = $P(T ; a, \lambda(T))$ using \eqnref{modulated-gamma}
    \State Update $a$ and  $p$ with Metropolis-Hastings and \eqnref{modulated-gamma}
\end{algorithmic}
\end{algorithm}

\section{Related Work}
\label{sec:prior-work}

\begin{figure*}[!tb]
  \begin{subfigure}[]{.33\linewidth}
    \centering \includegraphics[width=.95\linewidth]{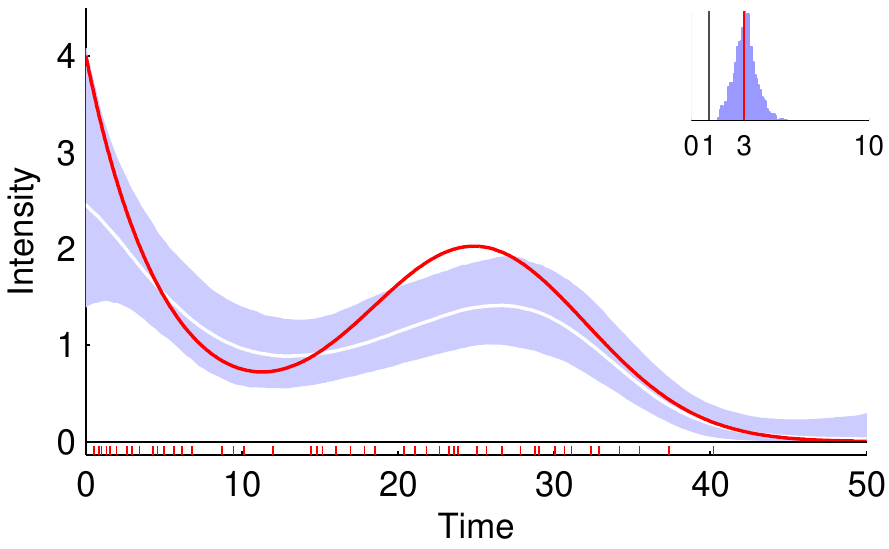}
  \end{subfigure}
  \begin{subfigure}[]{.33\linewidth}
    \centering \includegraphics[width=.95\linewidth]{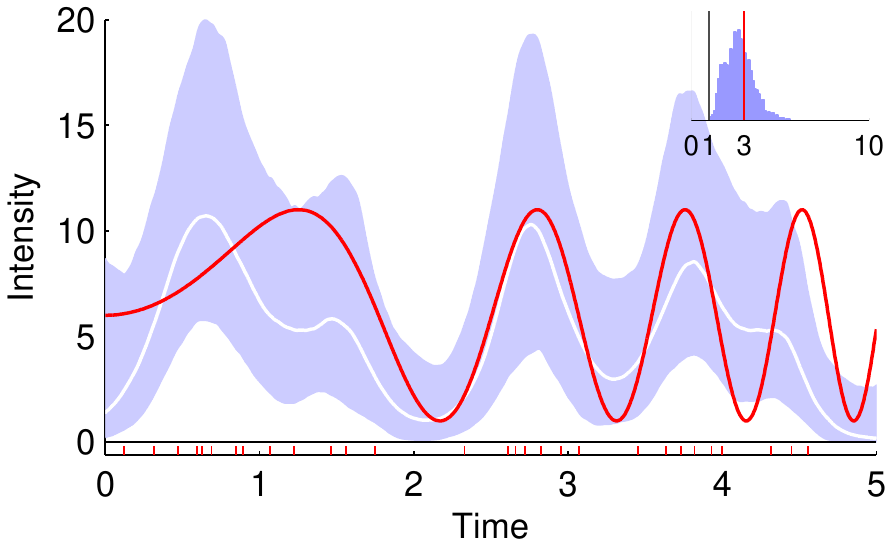}
  \end{subfigure}
  \begin{subfigure}[]{.33\linewidth}
    \centering \includegraphics[width=.95\linewidth]{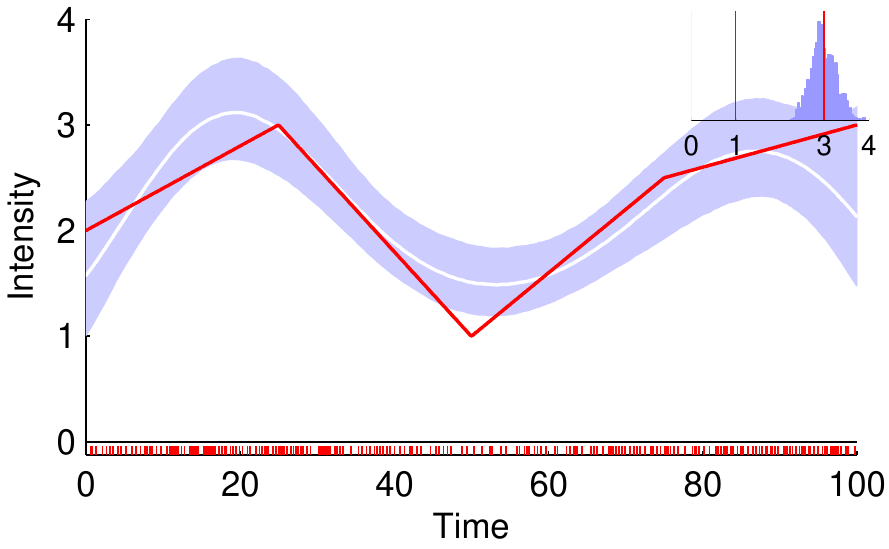}
  \end{subfigure}
  \begin{subfigure}[]{.33\linewidth}
    \centering \includegraphics[width=.95\linewidth]{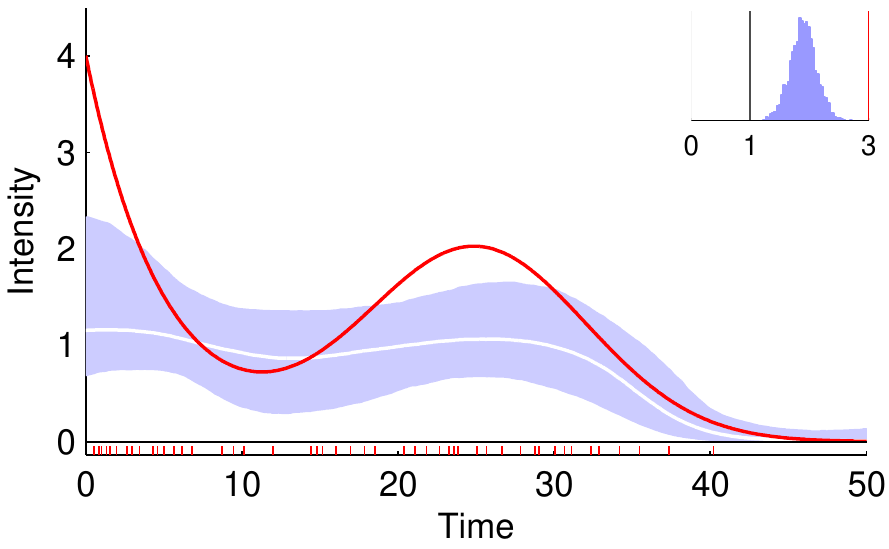}
  \end{subfigure}
  \begin{subfigure}[]{.33\linewidth}
    \centering \includegraphics[width=.95\linewidth]{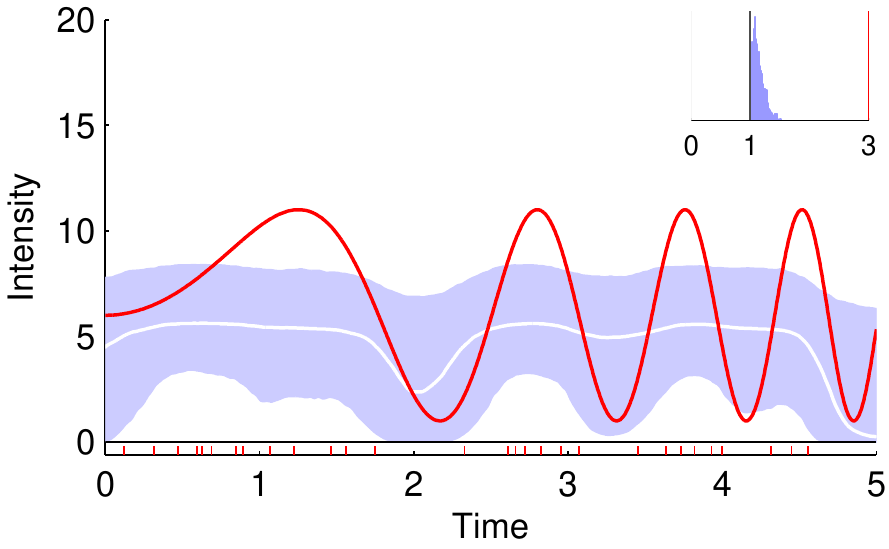}
  \end{subfigure}
  \begin{subfigure}[]{.33\linewidth}
    \centering \includegraphics[width=.95\linewidth]{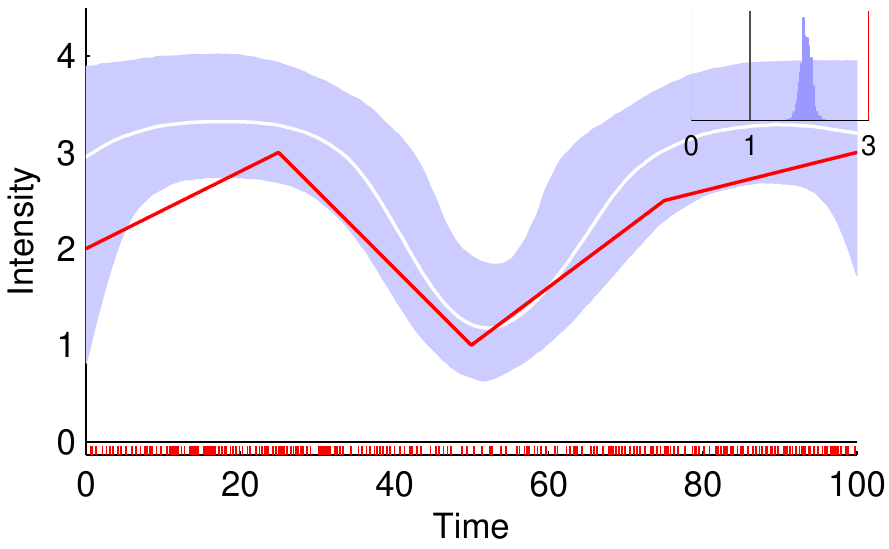}
  \end{subfigure}
  \caption{Our direct method (top) is more accurate than thinning (bottom) on parametric intensity
    functions $\lambda_1$ to $\lambda_3$ (left to right). Red line: true normalized intensity
    function $\lambda(t)/a$; White line: mean inferred normalized intensity function; Blue region:
    95\% confidence interval. Inset: inferred distribution of the gamma shape parameter $a$, with
    the true value marked in red. Grey bar at $a = 1$ for reference.}
  \label{fig:parametric-synthetic}
\end{figure*}

There is a growing literature on finding patterns among clinical variables such as laboratory tests
that have both a timestamp and a value \citep[see][and its references for examples]{Lasko2013}, but
we are not aware of any existing work exploring unsupervised, data-driven abstractions of the purely
time-domain clinical event streams that we address here.

There is much prior work on methods similar to ours that infer intensity functions for modulated
renewal processes. The main distinction between these methods lies in the way they handle the form
and integration of the intensity function $\lambda(t)$. Approaches include using kernel-smoothing
\citep{Ramlau-Hansen1983}, using parametric intensity functions \citep{Lewis1972a}, using
discretized bins within which the intensity is considered constant \citep{Moeller1998,
  Cunningham2008}, or using a form of rejection sampling called \emph{thinning} \citep{Adams2009a,
  Rao2011} that avoids the integration altogether.

The binned time approach is straightforward, and we share its use of Gaussian processes for the log
intensity function. However, there is an inherent information loss in the piecewise-constant
intensity function approximation. Moreover, its computational complexity is cubic with the number of
bins in the period of observation. For our data, with events at 1-day or finer time resolution over
up to a 15 year observation period, this method is prohibitively inefficient. A variant of this
approach that uses variable-sized bins \citep{Gunawardana2011, Parikh2012} has been applied to
medical data \citep{Weiss2013}. This variant is very efficient, but is restricted to a Poisson
process (fixed $a = 1$), and the inferred intensity functions are neither intended to nor
particularly well suited to forming an accurate abstraction over the raw discrete events.

Thinning is a clever method, but it is limited by the requirement of a bounded hazard function,
which prevents it from being used with bursty gamma processes (which have a hazard function
unbounded at zero). One thinning method has also adopted the use of Gaussian processes
\citep{Rao2011}, but is much less efficient than our algorithm, with time complexity cubic in the
number of events that would occur if the maximum event intensity were constant over the entire
observation time span. For event streams with a small dynamic range of intensities, this is not a
big issue, but our medical data sequences can have a dynamic range of several orders of magnitude.

Our method therefore has efficiency and flexibility advantages over existing methods, and we will
demonstrate in the experiments that it also has accuracy advantages.

\section{Experiments}
\label{sec:experiment}

In these experiments, we will refer to our inference method as the \emph{direct method} because it
uses direct numerical integration, as opposed to thinning, which avoids computing the integral at all.

We tested the ability of both methods to recover known intensity functions and shape parameters from
synthetic data. We then used the direct method to infer latent intensity functions from streams of
clinical events.

\subsection{Synthetic Data}
\label{sec:experiment-synthetic}

\begin{figure*}[!tb]
\centering
  \begin{subfigure}{.49\linewidth}
    \centering \includegraphics[width=.95\linewidth]{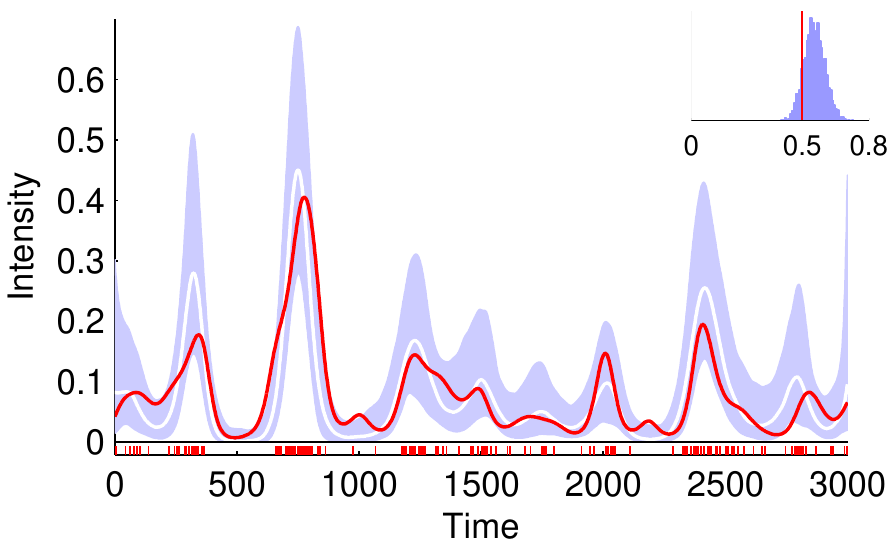}
  \end{subfigure}
  \begin{subfigure}{.49\linewidth}
    \centering \includegraphics[width=.95\linewidth]{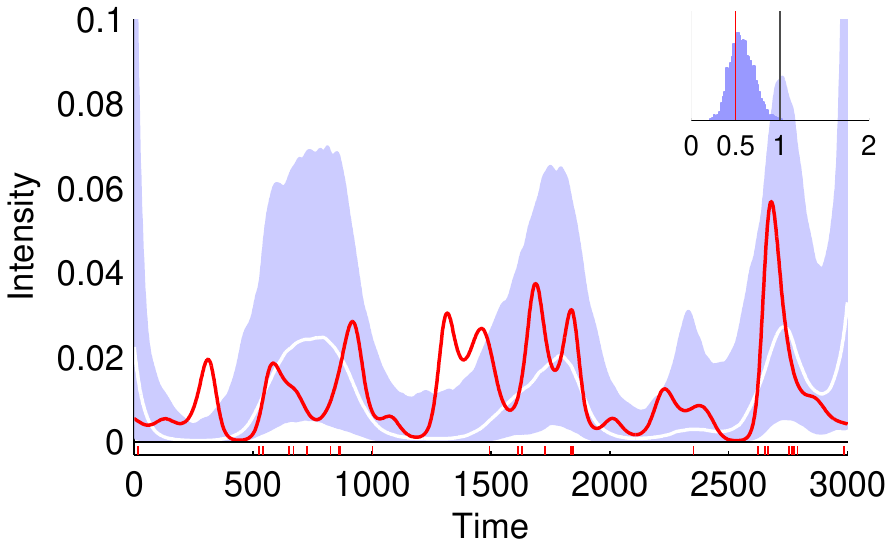}
  \end{subfigure}
  \caption{Accurate recovery of intensity function and parameters under conditions that would be
    prohibitive for any other method of which we are aware. Left panel presents results for high
    intensities and many events, right panel for low intensities and few events. While
    there is insufficient evidence in the right panel to recover the true intensity, the inferred
    intensity is reasonable given the evidence, and the inferred confidence intervals are accurate
    in that the true intensity is about 95\% contained within them. Legend as in Figure
    \protect\ref{fig:parametric-synthetic}.}
  \label{fig:synthetic}
\end{figure*}

Our first experiments were with the three parametric intensity functions below, carefully following
\citet{Adams2009a} and \citet{Rao2011}. We generated all data using the warping model described in
\secref{mrp}, with shape parameter $a = 3$.

\begin{enumerate}
\item $\lambda_1(t)/a = 2e^{-t/15} + e^{-((t - 25) / 10)^2}$ over the interval ${[0, 50]}$, 48 events.
\item $\lambda_2(t)/a = 5 \sin(t^2) + 6$ on $[0, 5]$, 29 events.
\item $\lambda_3(t)/a$ is the piecewise linear curve shown in Figure \ref{fig:parametric-synthetic},
  on the interval $[0, 100]$, 230 events.
\end{enumerate}
We express these as normalized intensities $\lambda(t)/a$, which have units of ``expected number of
events per unit time'', because they are more interpretable than the raw intensities and they are
comparable to the previous work done using Poisson processes, where $a = 1$.

We compared the direct method to thinning on these datasets. \citet{Adams2009a} compared thinning to
the kernel smoothing and binned time methods (all assuming a Poisson process), and \citet{Rao2011}
compared thinning to binned time, assuming a gamma process with constrained $a > 1$. Both found
thinning to be at least as accurate as the other methods in most tests.

We computed the RMS error of the true vs.\ the median normalized inferred intensity, the log probability of the data given
the model, and the inference run time under 1000 burn-in and 5000 inference MCMC iterations.

On these datasets the direct method was more accurate than thinning for the recovery of both the
intensity function and the shape parameter, and more efficient by up to two orders of magnitude
(Figure \ref{fig:parametric-synthetic} and Table \ref{tab:parametric-synthetic}). The results for
thinning are consistent with those previously reported \citep{Rao2011}.

\begin{table}[h]
  \caption{Performance on Synthetic Data. RMS: root-mean-squared error; LP: log probability of data
    given the model; RT: run time in seconds. Best results for each measure are bolded.}
\label{tab:parametric-synthetic}
\begin{center}
\begin{tabular}{lcccccc}
  \hline
  & \multicolumn{3}{c|}{Direct} &\multicolumn{3}{c}{Thinning} \\
  & \multicolumn{1}{c}{RMS} & \multicolumn{1}{c}{LP} & \multicolumn{1}{c|}{RT}
  & \multicolumn{1}{c}{RMS} & \multicolumn{1}{c}{LP} & \multicolumn{1}{c}{RT}\\
  \hline
  $\lambda_1$  & $\mathbf{0.37}$ & $\mathbf{+12.1}$& $\mathbf{453}$ & $0.66$ & $-62.7$ & $4816$\\
  $\lambda_2$  & $\mathbf{3.1}$ & $\mathbf{-228}$& $\mathbf{511}$ & $3.4$ & $-333$ & $1129$ \\
  $\lambda_3$  & $\mathbf{0.25}$ & $\mathbf{+0.293}$ & $\mathbf{385}$ & $0.53$ & $-82.2$ & $41291$\\
\end{tabular}
\end{center}
\end{table}

Additionally, the confidence intervals from the direct method are subjectively more accurate than
from the thinning method. (That is, the 95\% confidence intervals from the direct method
contain the true function for about 95\% of its length in each case). This is particularly important
in the case of small numbers of events.

As might be expected, we found the results for $\lambda_2(t)$ to be sensitive to the prior
distribution on $l$, given the small amount of evidence available for the inference. Following
\citet{Adams2009a} and \citet{Rao2011}, we used a log-normal prior with a mode near $l = 0.2$, tuned
slightly for each method to achieve the best results. We also allowed thinning to use a log-normal
prior with appropriate modes for $\lambda_1(t)$ and $\lambda_3(t)$, to follow precedent in the
previous work, although it may have conferred a small advantage to thinning. We used the weaker
exponential prior on those datasets for the direct method.

Our next experiments were on synthetic data generated to resemble our medical data. We tested
several configurations over wide ranges of parameters, including some that were
not amenable to any known existing approach (such as the combination of $a < 1$, high dynamic range
of intensity, and high ratio of observation period to event resolution, Figure \ref{fig:synthetic}).
The inferred intensities and gamma parameters were consistently accurate. Estimates of the
confidence intervals were also accurate.

\subsection{Clinical Data}
\label{sec:experiment-clinical}

\begin{figure*}[htb]
\centering
  \begin{subfigure}{.49\linewidth}
    \centering \includegraphics[width=.95\linewidth]{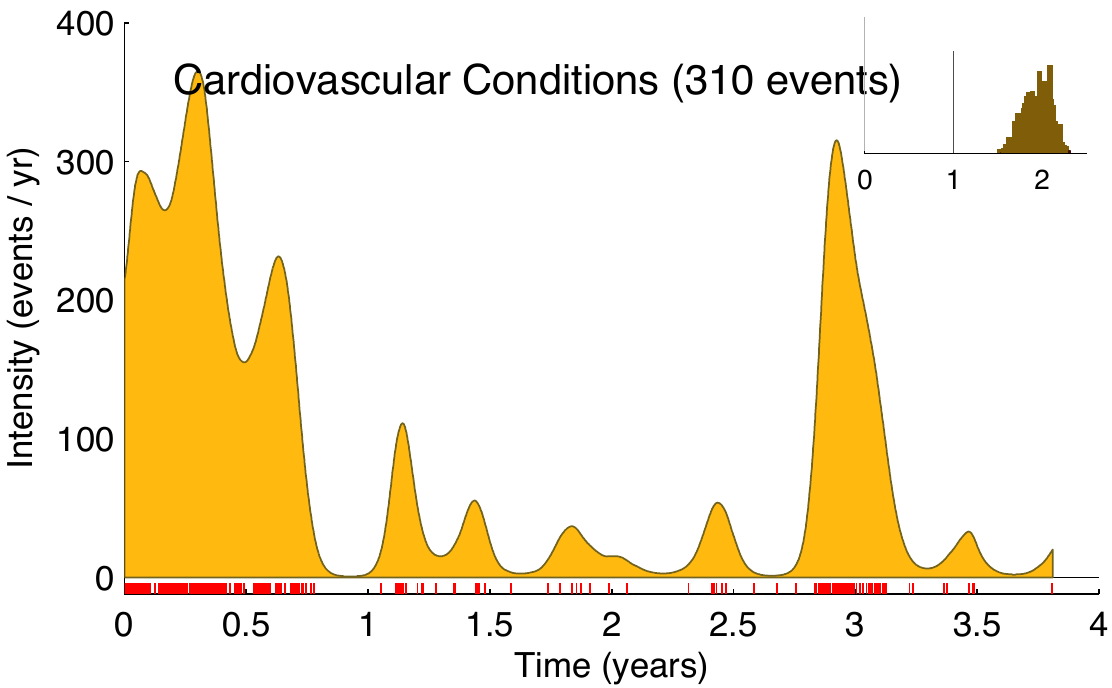}
    \caption{} \label{fig:CV}
  \end{subfigure}
  \begin{subfigure}{.49\linewidth}
    \centering \includegraphics[width=.95\linewidth]{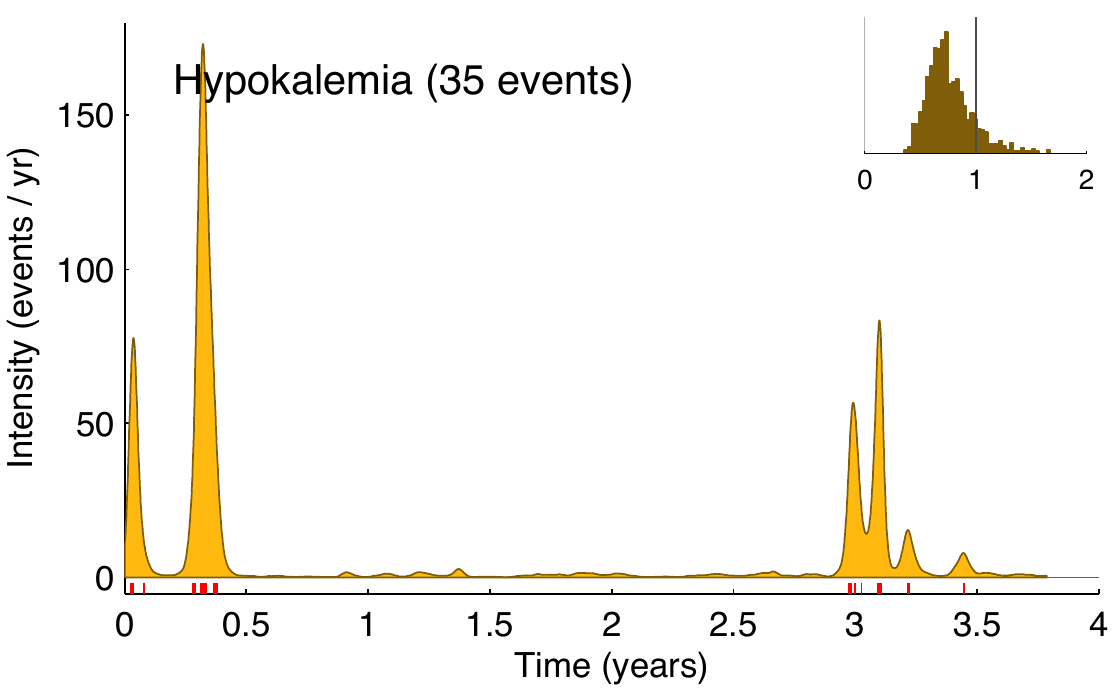}
    \caption{}\label{fig:hypokalemia}
  \end{subfigure}
  \caption{Example inferred intensities for a large class of conditions (\subref{fig:CV}) and for a specific
    disease (\subref{fig:hypokalemia}) in one patient's record. Inset: inferred posterior
    distribution of the gamma shape parameter, with marker at $a = 1$ to aid
    interpretation. For clarity, confidence intervals on intensity are not shown.}
  \label{fig:clinical}
\end{figure*}

\begin{figure*}[ptb]
    \centering \includegraphics[width=\linewidth]{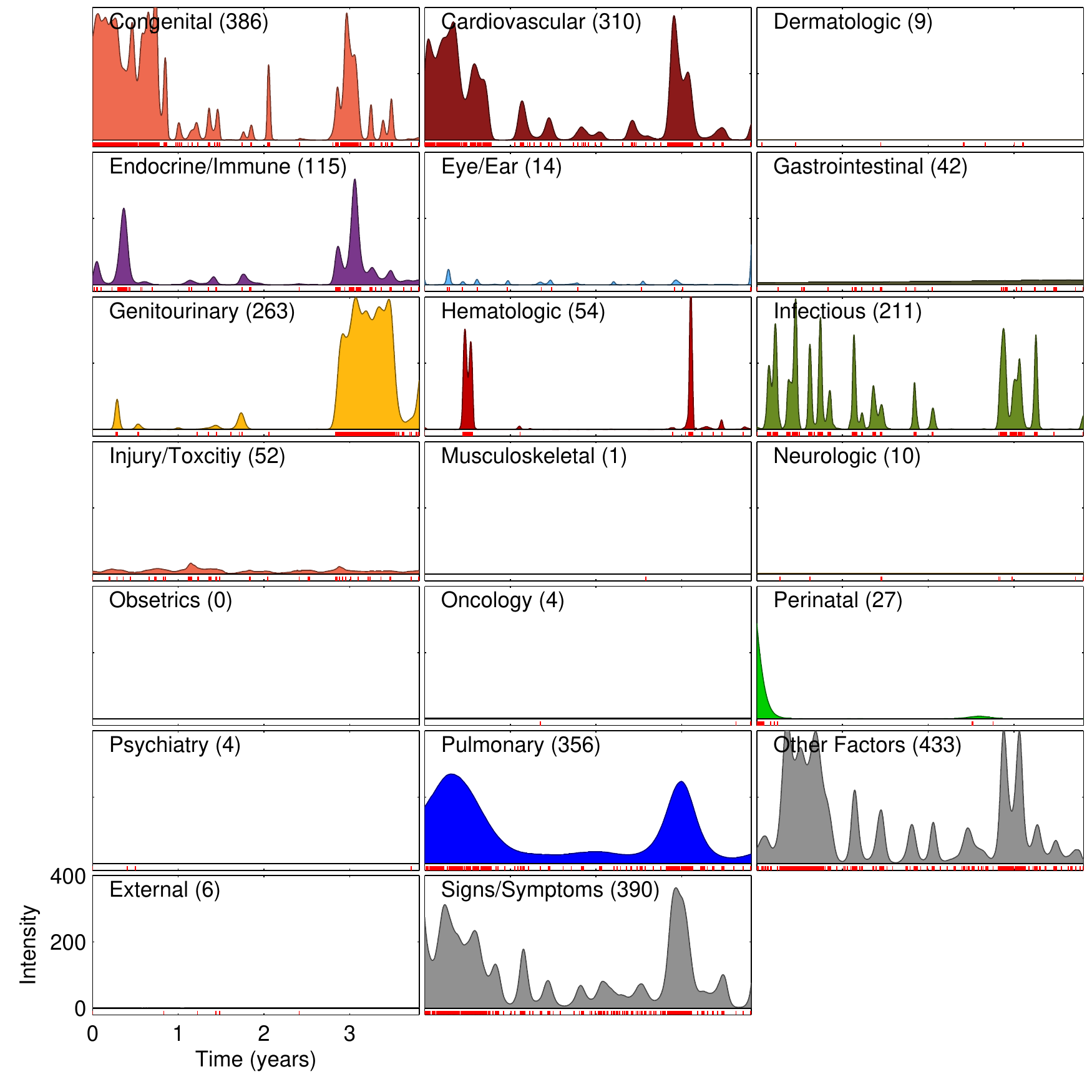}
    \caption{Inferred intensities for all top-level ICD-9 divisions of a very complicated patient's
      record. Such a display may be clinically useful for getting a quick, broad understanding of a
      patient's medical history, including quickly grasping which conditions have \emph{not} been
      diagnosed or treated. Numbers in parentheses: total number of events in each division.  For
      clarity, confidence intervals are not shown.}
  \label{fig:summary}
\end{figure*}

Lastly, we applied the direct method to sequences of billing codes representing clinical
events. After obtaining IRB approval, we extracted all \mbox{ICD-9} codes from five patient records
with the greatest number of such codes in
the deidentified mirror of
our institution's Electronic Medical Record. We arranged the codes from each patient record as
streams of events grouped at both the top level of the ICD-9 disease hierarchy (groups of broadly
related conditions), and at the level of the individual disease.

For the streams of grouped events, we included an event if its associated ICD-9 code fell within the
range of the given top-level division. For example, any \mbox{ICD-9} event with a code in the range
[390 -- 459.81] was considered a \emph{Cardiovascular} event. While intensity functions are only
strictly additive for Poisson processes, we still find the curves of grouped events to be
informative.

We inferred intensity functions for each of these event streams (Figures \ref{fig:clinical} and
\ref{fig:summary}). Each curve was generated using 2000 burn-in and 2000 inference iterations in
about three minutes using unoptimized MATLAB code on a single desktop CPU. The results of both types
have good clinical face validity.

The set of top-level intensity functions make clear that there is much underlying structure in these
originally irregular and asynchronous medical events that can now be investigated with standard
learning methods \figref{summary}. For example, there are obvious dependencies in this patient
between Congenital, Cardiovascular, and Pulmonary conditions, and then a dependence of those with a
Genitourinary condition emerges around year 3. Such structure can be investigated at both the
patient and the population level using these abstractions.

\section{Discussion}
We have made two contributions with this paper. First, we presented a direct numeric method to infer
a distribution of continuous intensity functions from a set of episodic, irregular, and discrete
events. This direct method has increased efficiency, flexibility, and accuracy compared to the best
prior method. Second, we presented results using the direct method to infer a continuous function
density as an abstraction over episodic clinical events, for the purposes of transforming the raw
event data into a form more amenable to standard machine learning algorithms.

The clinical interpretation of these intensity functions is that increased intensity represents
increased frequency of contact with the healthcare system, which usually means increased
\emph{instability} of that condition. In some cases, it may also mean increased \emph{severity} of
the condition, but not always. If a condition acutely increases in severity, this represents an
instability and will probably generate a contact event. On the other hand, if a condition is severe
but stably so, it may or may not require high-frequency medical contact.

A method to construct similar curves from observations with both a time and a continuous value has
been previously reported \citep{Lasko2013}, and we have presented here a method to construct them
from observations with a time plus a categorical label. These two data types represent the majority
of the information in a patient record (if we consider words and concepts in narrative text to be
categorical variables), and opens up many possibilities for finding meaningful patterns in large
medical datasets.

The practical motivation for this work is that once we have the continuous function densities, we
can use them as inputs to a learning problem in the time domain (such as identifying trajectories
that may be characteristic of a particular disease), or by aligning many such curves in time and
looking for useful patterns in their cross-sections (which to our knowledge has not yet been
reported).

We discovered incidentally that a presentation such as Figure \ref{fig:summary} appears to be a
promising representation for efficiently summarizing a complicated patient's medical history and
communicating that broad summary to a clinician. The presentation could allow drilling-down to the
intensity plots of the specific component conditions and then to the raw source data. (The usual
method of manually paging through the often massive chart of a patient to get this information can
be a tedious and frustrating process.)

One could also imagine presenting the curves of not the raw ICD-9 divisions, but the inferred latent
factors underlying them, and drilling down into the rich combinations of test results, medications,
narrative text, and discrete billing events that comprise those latent factors.  We plan to
investigate these possibilities in future work.

\subsubsection*{Acknowledgements}
This work was funded by grants from the Edward Mallinckrodt, Jr. Foundation and the National
Institutes of Health 1R21LM011664-01. Clinical data was provided by the Vanderbilt Synthetic
Derivative, which is supported by institutional funding and by the Vanderbilt CTSA grant ULTR000445.

\end{document}